\documentclass{article}
\usepackage{ijcai24}

\usepackage{times}
\usepackage{soul}
\usepackage{url}
\usepackage[hidelinks]{hyperref}
\usepackage[utf8]{inputenc}
\usepackage[small]{caption}
\usepackage{graphicx}
\usepackage{amsthm}
\usepackage{booktabs}
\usepackage{algorithm}
\usepackage{algorithmic}
\usepackage[switch]{lineno}

\usepackage{amsmath,amsfonts}
\usepackage{array}
\usepackage{textcomp}
\usepackage{stfloats}
\usepackage{verbatim}
\usepackage{subfloat}
\usepackage{subfig}
\usepackage{utfsym}
\usepackage{placeins} 

\usepackage{multirow}
\usepackage{color}
\usepackage{xcolor}
\usepackage{makecell}
\usepackage{siunitx}

\title{FastScene: Text-Driven Fast 3D Indoor Scene Generation \\
via Panoramic Gaussian Splatting}

\author{
Yikun Ma$^1$
\and
Dandan Zhan$^1$\and
Zhi Jin$^{1,2}$ \footnote{Corresponding author}
\affiliations
$^1$Sun Yat-sen University\\
$^2$Guangdong Provincial Key Laboratory of Fire
Science and Intelligent Emergency Technology\\
\emails
\{mayk25, zhandd3\}@mail2.sysu.edu.cn,
jinzh26@mail.sysu.edu.cn
}

\begin{document}

\maketitle

\begin{abstract}
Text-driven 3D indoor scene generation holds broad applications, 
ranging from gaming and smart homes to AR/VR applications. 
Fast and high-fidelity scene generation is paramount for ensuring user-friendly experiences. 
However, existing methods are characterized by lengthy generation processes 
or necessitate the intricate manual specification of motion parameters, which introduces inconvenience for users. 
Furthermore, these methods often rely on narrow-field viewpoint iterative generations, 
compromising global consistency and overall scene quality. 
To address these issues, we propose FastScene, a framework for fast and higher-quality 3D scene generation, 
while maintaining the scene consistency. 
Specifically, given a text prompt, we generate a panorama and estimate its depth, 
since the panorama encompasses information about the entire scene and exhibits explicit geometric constraints. 
To obtain high-quality novel views, we introduce the Coarse View Synthesis (CVS) and Progressive Novel View Inpainting (PNVI) strategies, 
ensuring both scene consistency and view quality. Subsequently, we utilize Multi-View Projection (MVP) to form perspective views, 
and apply 3D Gaussian Splatting (3DGS) for scene reconstruction. 
Comprehensive experiments demonstrate FastScene surpasses other methods in both generation speed and quality with better scene consistency. 
Notably, guided only by a text prompt, FastScene can generate a 3D scene within a mere 15 minutes, 
which is at least one hour faster than state-of-the-art methods,
making it a paradigm for user-friendly scene generation. 
 
\end{abstract}

\section{Introduction}

3D models have a wide range of applications in video production, gaming, AR/VR, and other fields. 
However, generating high-quality 3D models typically requires professional designers to 
utilize specialized software with a considerable amount of time, 
which is inconvenient for those seeking fast 3D model generation. 
The development of generative models makes 
Text-to-3D object generation \cite{poole2022dreamfusion}, \cite{lin2023magic3d} possible and impressive. 
However, the generation of 3D scenes still presents significant challenges, 
requiring large-scale scene reconstruction, multi-view images, and the assurance of scene realism and consistency.

Recently, some works attempt to tackle the 3D scene generation challenges. 
Set-the-Scene \cite{cohen2023set} applies global-local training from text prompts and 3D object proxies, 
while generating controllable scenes. 
However, the quality and resolution of the generated scenes are unsatisfactory due to the lack of corresponding geometry. 
SceneScape \cite{fridman2023scenescape} generates long-range views, producing diverse styles.  
However, its view quality decreases over time due to the inpainting and depth estimation error accumulation. 
Text2Room \cite{hoellein2023text2room} and Text2NeRF \cite{zhang2024text2nerf} gradually generate perspective novel views. 
Nevertheless, their incremental local operations hardly ensure scene consistency and coherence. 
Ctrl-Room \cite{fang2023ctrl} fine-tunes ControlNet \cite{zhang2023adding} for editable panorama generation, 
and then performs mesh reconstruction. 
However, Ctrl-Room tends to flatten the 3D model with limited scene quality, 
since it hardly generates multi-view images.

As one of the 3D representation techniques, the radiance fields methods, 
exemplified by Neural Radiance Fields (NeRF) \cite{mildenhall2020nerf}, 
have made significant breakthroughs. 
Since most NeRF-based methods suffer from slow rendering speed \cite{mildenhall2020nerf}, \cite{barron2022mip}, 
rendering process acceleration becomes an important issue. 
Recently, 3D Gaussian Splatting (3DGS) \cite{kerbl20233d} 
has achieved success in the rendering speed with high-quality. 
However, the typical 3DGS only takes regular images as the input.  
It faces challenges when handling panoramas, 
which are difficult to handle with existing Structure-from-Motion (SFM) \cite{snavely2006photo} methods.

\begin{figure*}[t]
    \centering
    
    \includegraphics[width=0.99\linewidth]{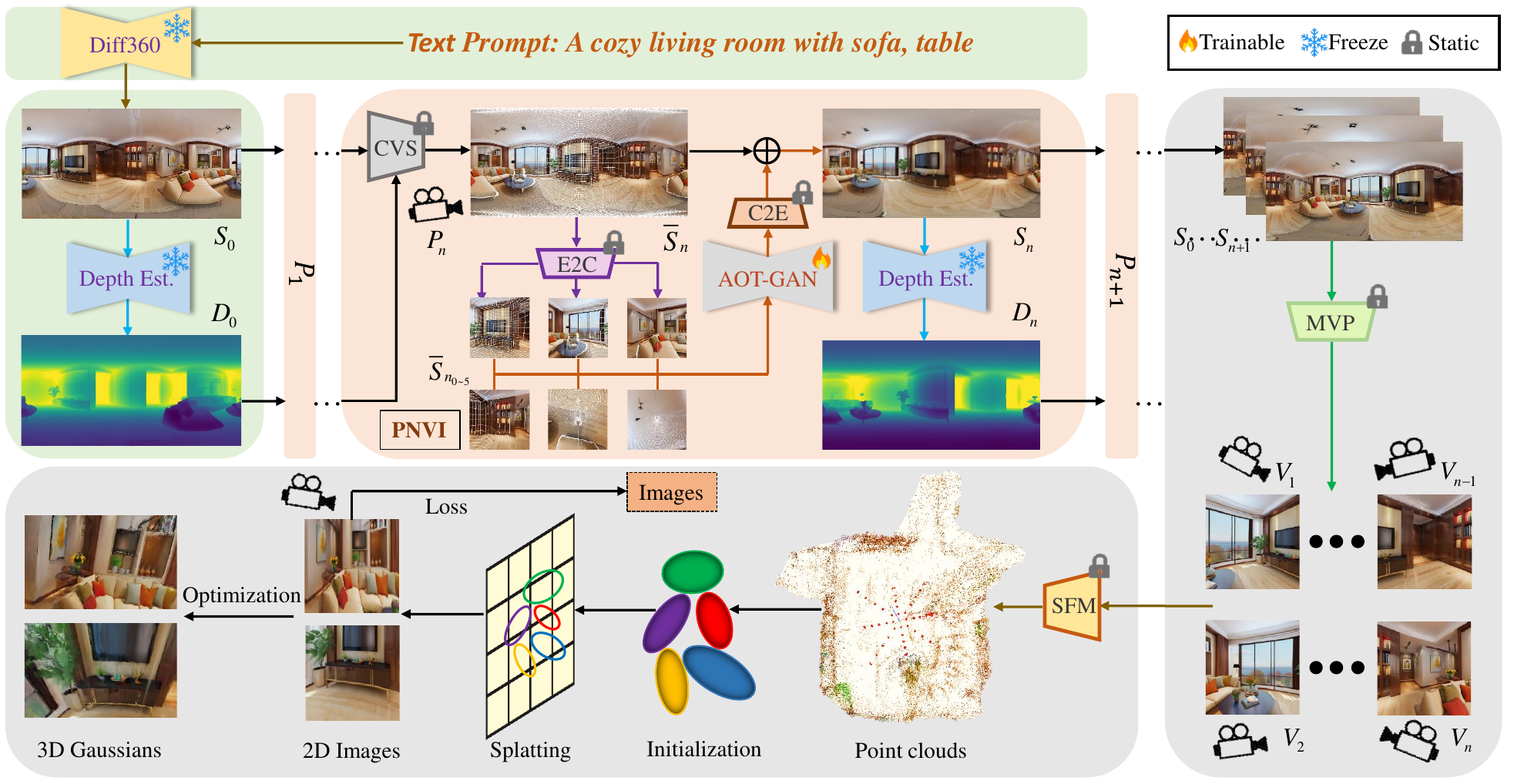}

    \caption{The framework of our FastScene. Given the text prompt, 
    we first generate a panorama and estimate its depth. 
    Then, we iteratively generate multi-view panoramas through PNVI. 
    We introduce MVP for perspective projection and use 3DGS for scene reconstruction.}
    \label{fig:pipeline}
    \vspace{-0.5cm}
\end{figure*}

To address the above issues, we propose a novel Text-to-3D scene framework, 
called FastScene, which aims at fast generating consistent and authentic scenes with high-quality. 
As shown in Figure \ref{fig:pipeline}, our approach primarily comprises three stages. 
\textbf{1)} In the first stage, given a text prompt, we generate a panorama by utilizing the pre-trained Diffusion360 \cite{feng2023diffusion360}. 
Panorama is selected due to its ability to capture the global information and exhibit explicit geometric constraints, 
which is advantageous in overcoming the scene inconsistency issue of perspective view. 
Then, we adopt EGformer \cite{yun2023egformer} for panorama depth estimation. 
\textbf{2)} In the second stage, we propose Coarse View Synthesis (CVS) to generate novel panoramic views with holes for specific camera poses. 
Since large-distance novel views generation results in numerous holes, which is not conducive to inpainting, 
we propose Progressive Novel View Inpainting (PNVI) to gradually fill the holes within a small distance. 
Nevertheless, we experimentally find that caused by cumulative distortion errors, 
directly inpainting panorama usually results in edge blurring and distortion. 
Instead, we propose to perform inpainting in cubemap, and utilize Cube-to-Equirectangular (C2E) to obtain the corresponding inpainted panorama. 
We then replace non-hole pixels in the inpainted panorama with their original values. 
Furthermore, we synthesize a dataset that aligns with our hole distribution, and retrain AOT-GAN \cite{zeng2022aggregated} for inpainting. 
\textbf{3)} After acquiring multi-view panoramas, we employ 3DGS for fast scene reconstruction. 
However, the original COLMAP \cite{schonberger2016structure} only supports perspective views, 
making it challenging to obtain point clouds from panoramas. 
Therefore, we introduce Multi-View Projection (MVP), which divides the panorama into perspective views, 
enabling feeding into 3DGS for reconstruction. 
MVP as a plug-and-play module can be easily applied without requiring additional computational resources. 
Extensive experiments validate that our method can fast generate high-quality 3D scenes while ensuring scene consistency.

Our contributions can be summarized as follows:
\begin{itemize}
    \item [1)]
    We propose a novel Text-to-3D indoor scene framework FastScene, 
    enabling fast and high-quality scene generation, while ensuring scene consistency.  
    Additionally, given the text prompt, there is no need to pre-design complex camera parameters or motion trajectories, 
    which makes FastScene a user-friendly scene generation paradigm. 
    \item [2)]
    We propose a novel panoramic view synthesis method PNVI, which adopts CVS to generate novel views with holes, 
    and performs precision-controllable progressive inpainting to generate refined views. 
    Additionally, to improve the inpainting quality, we synthesize a large-scale distribution-based spherical mask dataset. 
    \item [3)]
    To the best of our knowledge, we are the first to solve panoramic 3DGS from a single panorama, 
    and the proposed FastScene is highly adaptable to existing panoramic data for reconstruction.
\end{itemize}

The rest of this paper is organized as follows.
Section ~\ref{RW} briefly reviews the related works of this paper. 
Section ~\ref{Method} introduces the design details of the proposed FastScene. 
Section ~\ref{Exper} provides experimental results for comparisons and ablation study. 
Conclusions are summarized in Section ~\ref{conc}.

\section{Related Works}\label{RW}

\subsection{Text-Driven 3D scene Generation}

Recently, there has been considerable focus on 3D scene generation. 
\ Set-the-Scene \cite{cohen2023set} introduces an agent-based global-local framework to synthesize controllable 3D scenes, 
while enabling diverse scene editing options. 
However, it suffers from shortcomings in the quality and resolution of generation scenes without corresponding geometry. 
While SceneScape \cite{fridman2023scenescape} generates consistent views by introducing a pre-trained text-to-image model \cite{rombach2022high}, 
and possesses the capability to generate scenes in various styles. 
However, the view quality of SceneScape is reliant on geometric priors and diminishes over time due to the inpainting and depth error accumulation. 
More recently, both Text2Room \cite{hoellein2023text2room} and Text2NeRF \cite{zhang2024text2nerf} rely on incremental frameworks to generate new perspectives on a per-image basis. 
However, their incremental local operations can hardly guarantee scene consistency and coherence. 
Later on, Ctrl-Room \cite{fang2023ctrl} proposes to encode text input and convert scene code into a 3D bounding box for editing. 
Subsequently, it generates panoramas by fine-tuning ControlNet \cite{zhang2023adding}, 
and reconstructs mesh through Possion reconstruction \cite{kazhdan2006poisson} and MVS-texture \cite{waechter2014let}. 
However, Ctrl-Room struggles to generate high-quality 3D models, 
and tends to flatten the 3D model due to the limited number of generated views.

\subsection{Text-Driven Panorama Generation}
Unlike 2D images, panoramas cover the 360\textdegree \ $\times$ 180\textdegree \ field of view, 
which provides more 3D scene information. 
Text2Light \cite{chen2022text2light} synthesizes panorama images from text input via a multi-stage auto-regressive generative model. 
However, it ignores the boundary continuity of the panorama, resulting in an open-loop content. 
MVDiffusion \cite{tang2023mvdiffusion} generates high-resolution panoramas by fine-tuning a pre-trained text-to-image diffusion model. 
However, artifacts usually appear on the ``sky'' and ``floor'' views, 
which decreases the realism of generated scenes.
StitchDiffusion \cite{wang2024customizing} crops the left and right sides of the panorama to maintain the scene continuity. 
However, the cracks at the seams are still noticeable.
Diffusion360 \cite{feng2023diffusion360} proposes a circular blending strategy to maintain the geometry continuity, 
which generates high-resolution boundary-continuous panoramas. 

\subsection{Novel View Synthesis}
Novel view synthesis is a popular area of significant interest. 
Early methods rely on multi-view images and attempt to incorporate the knowledge 
from epipolar geometry to perform smooth interpolation between the different views \cite{chen1993view}, \cite{debevec1996modeling}. 
Some methods synthesize novel views by deep networks from a few images \cite{sajjadi2022scene}, \cite{mirzaei2023reference}.
In contrast, \cite{gu2023nerfdiff}, \cite{shen2023anything} allow for generating novel views from a single image. 

A significant breakthrough in novel view synthesis is NeRF \cite{mildenhall2020nerf} 
and its derivative works \cite{barron2021mip}, \cite{barron2022mip}, \cite{chen2023testnerf}. 
The rendering speed of most radiance-based methods is slow, 
accelerating rendering becomes an important but challenging problem, 
with representative works such as Instant-NGP \cite{muller2022instant} and 3DGS \cite{kerbl20233d}. 
Some NeRF-based works \cite{wang2023perf}, \cite{chen2024panogrf} attempt to synthesize panoramic novel views. 
However, since SFM struggles to handle panoramas due to its unique structure \cite{snavely2006photo}, 
it is difficult to utilize original 3DGS for panorama rendering.

\section{Method}\label{Method}

\subsection{Overview}
As shown in Figure \ref{fig:pipeline}, given a text prompt $P$, 
we first use Diffusion360 \cite{feng2023diffusion360} to generate the corresponding panorama $S_0$, 
and then employ EGformer \cite{yun2023egformer} to estimate the depth map $D_0$.  
Thereafter, given a new camera pose $P_n$, we perform CVS
to obtain the corrupted panorama $\overline{S}_n$ with holes. 
To fill these holes, we propose PNVI, 
which gradually inpaints perspective cubemap views $\overline{S_{n_i}} (i = 0, 1, ... 5)$ rather than directly inpaints the panorama.
Subsequently, these clean cubemap images are reprojected equidistantly to obtain the clean panorama.  
We then replace non-hole pixels in the inpainted panorama with their original values to obtain the novel panorama $S_n$. 
Similarly, iterating PNVI multiple times results in numerous novel clean panoramic views. 
As COLMAP \cite{schonberger2016structure} does not support panoramic inputs, 
we employ MVP to generate the corresponding perspective views, 
followed by 3DGS to implement the 3D scene reconstruction.

\subsection{Text-Driven Panorama Generation and CVS}

\begin{figure}[b]
    \centering
    \vspace{-0.3cm}
    \includegraphics[width=0.65\linewidth]{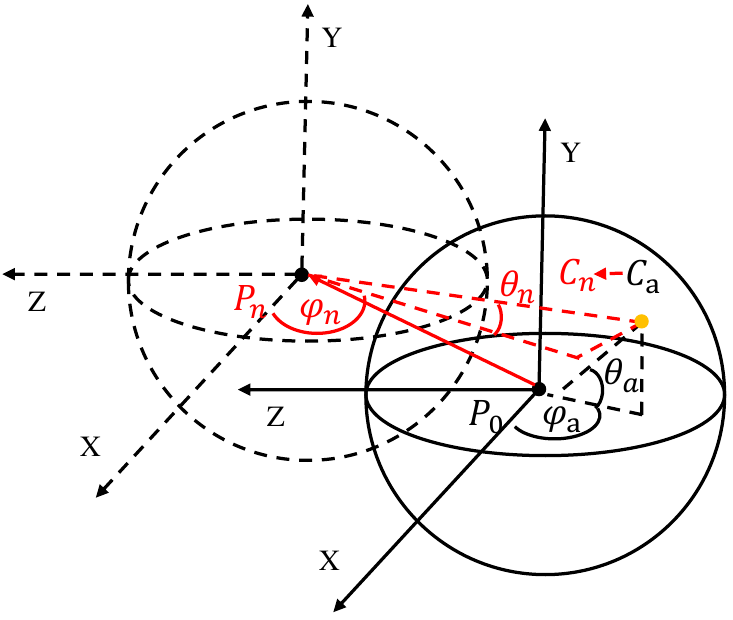}
    \caption{Given a new camera pose $P_n$, 
    the calculation for movement in spherical coordinates.}
    \label{qiu}
    \vspace{-0.2cm}
\end{figure}

Compared to perspective views, 
a key geometric characteristic of panorama is the continuity of the boundaries.
Additionally, the panorama encompasses information about the entire scene and exhibits explicit geometric constraints, 
which is beneficial for our subsequent processing. 
Thus, we utilize Diffusion360 \cite{feng2023diffusion360} for text-to-panorama generation, 
which adopts the blending strategy to maintain the geometry continuity. 
After that, we estimate the depth map using EGformer \cite{yun2023egformer} 
to capture the spatial information of the scene. 
Then, we propose CVS to obtain a new panoramic view under a given camera pose, as shown in Figure \ref{qiu}. 
According to the theory of equidistant projection on the spherical panorama, 
we can project a 2D image of size 1024 $\times$ 512 onto a sphere, 
where the latitude range is $180^\circ$ and the longitude range is $360^\circ$. 
The calculation for the latitude angle $\theta_a$ and longitude angle $\phi_a$ are as follows: 
\begin{eqnarray}
    & \theta_a = \frac{\pi y_a}{H} ,   \\
    & \phi_a =  \frac{2\pi x_a}{W} ,
\end{eqnarray}
where $x_a$ and $y_a$ represent the image coordinates of coordinate system $a$, 
while $W$ and $H$ represent the width and height of the panorama, respectively. 
We then utilize the triangle transformation to obtain the spherical basis coordinates: 
\begin{eqnarray}
    a_x  &=& \cos^{\theta_a} \cdot \cos^{\phi_a} ,  \\
    a_y  &=& \sin^{\theta_a} ,   \\
    a_z  &=& - \ \cos^{\theta_a} \cdot \sin^{\phi_a} , 
\end{eqnarray} 
afterward, we multiply the depth $d$ by the 3D coordinates $a_x,a_y,a_z$ 
to initial the spherical coordinates $C_a$: 
\begin{eqnarray}
    C_a = (d\cdot a_x, d\cdot a_y, d\cdot a_z) .
\end{eqnarray}

Given a new camera pose $P_n$, we take it as the origin of the new spherical coordinate system $n$, 
and subtract the original coordinates $C_a$ from the new origin $P_n$ to get the new spherical coordinates: 
\begin{eqnarray}
    C_n = (n_x, n_y, n_z) = \frac{C_a - P_n}{|C_a - P_n|} .
\end{eqnarray}

Then, we reproject the coordinates $C_n$ to the new coordinate system $n$: 
\begin{eqnarray}
        \theta_n &=& \arctan \frac{n_y}{\sqrt{{n_x}^2 + {n_z}^2}} , \\
        \phi_n &=&  \arctan \frac{-n_z}{n_x} , \\
        x_n &=& \frac{\phi_n}{2 \pi}W , \\
        y_n &=& \frac{\theta_n}{\pi}H ,
\end{eqnarray}
where $\theta_n$ and $\phi_n$ denote the latitude and longitude of the novel view, 
and $x_n$ and $y_n$ represent the image coordinates of coordinate system $n$. 

We summarize equations (1) to (11) as a mapping $F$ from $(x_a, y_a)$ to $(x_n, y_n)$: 
\begin{eqnarray}
    (x_n, y_n) = F (x_a, y_a) .
\end{eqnarray}

Therefore, we only need to determine if the mapped pixels $(x_n, y_n)$ lie within the panorama.  
If they are inside, we keep the normal RGB values, 
otherwise we set them as holes with a value 255:  
\begin{eqnarray}
    \overline{S}_n = \begin{cases}
        normal, & \text{if } (x_n \leq W, y_n \leq H, d_n > 0) \\
        255, & \text{otherwise}
    \end{cases}
\end{eqnarray}
Correspondingly, we can obtain the mask image $M_n$ 
(with value 0 for normal regions and 1 for unseen) for inpainting: 
\begin{equation}
    M_n = \begin{cases}
        0, & \text{if } (\overline{S}_n = normal) \\
        1, & \text{otherwise}
    \end{cases}
\end{equation}

\begin{figure}[!t]
    \centering
    \subfloat
    {
        \includegraphics[width=0.95\linewidth]{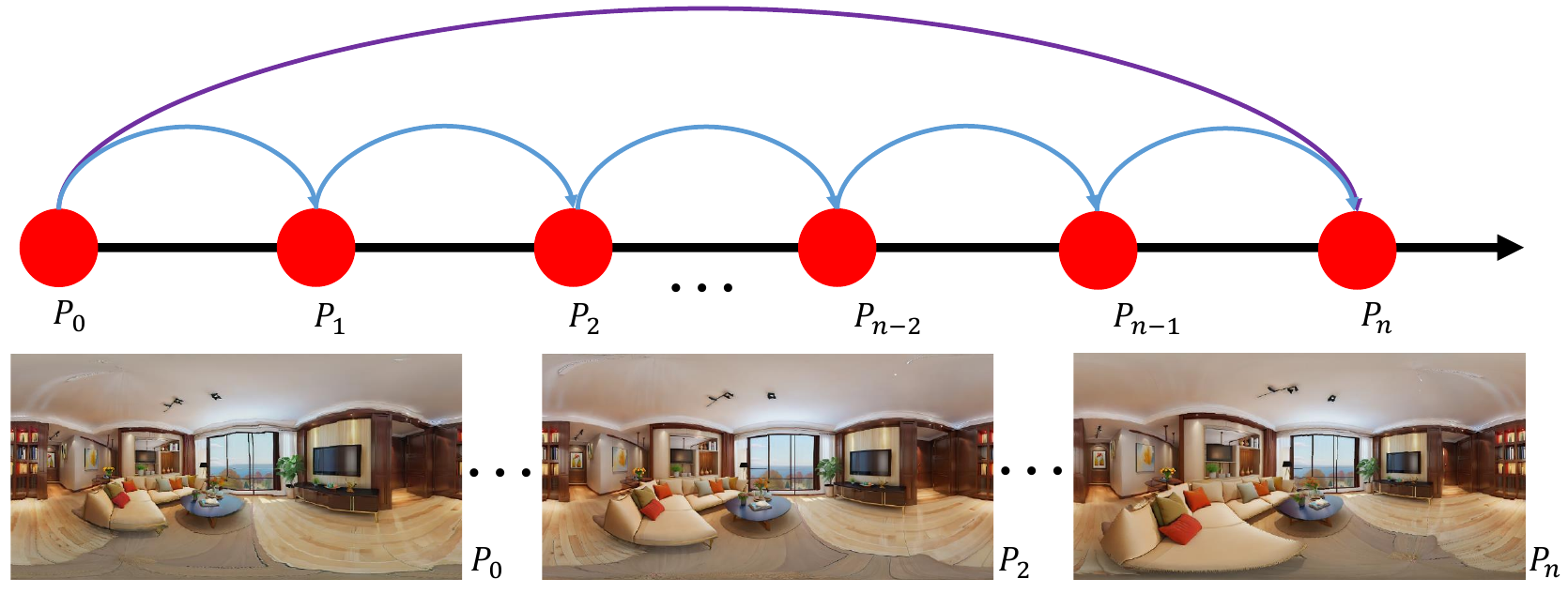}
    }
    \vspace{-0.2cm}
    \caption{Illustration of progressive inpainting and movement.}
    \label{Inpainting_flow}
    \vspace{-0.4cm}
\end{figure}

\subsection{Progressive Novel View Inpainting}
After CVS, we obtain multi-view panoramas with holes.  
To reconstruct the scene using 3DGS, we need to fill these holes. 
To this end, we propose the PNVI to obtain clean novel views. 
Due to the lack of indoor panoramic datasets with mask information to retrain the inpainting network, 
we construct a new dataset, as detailed in Section \ref{dataset}. 
We endeavored to conduct direct panorama inpainting, yet observed that with an increasing distance of movement, 
a plethora of spurious shadows manifested along the peripheries of the panorama. 
Therefore, E2C is utilized to obtain six cubemap images from one panorama, 
and cubemap inpainting is conducted using the retrained AOT-GAN \cite{zeng2022aggregated}. 
After that, C2E is utilized to form the panorama. 
Finally, we replace non-hole pixels in the inpainted panorama with their original values to obtain the novel panorama $S_n$.

However, when directly moving the camera to large poses, the hole-to-image area ratios become extensive, 
raising difficulties for inpainting, irrespective of the model training quality. 
To address the aforementioned issue, we propose a progressive inpainting mode, as shown in Figure \ref{Inpainting_flow}, 
which enables inpainting in large camera poses. 
Specifically, assuming we move the camera along the X-axis by a distance of 0.33 meters,  
the hole-to-image area ratio of the novel view image increases to 64.3$\%$, 
which means more than half of the images are with holes, as reported in Table \ref{tab:mask}. 
Therefore, we decide to divide the long distance into small moves (e.g., 0.02 meters per move) 
to relieve the long distance inpainting difficulty. In this way, the hole-to-image ratio is only 15$\%$ at each move.
By progressively moving from $P_0$ to $P_n$, we can obtain a clean view at the endpoint.
\begin{table}[htbp]
    \vspace{-.1cm}
    \setlength{\abovecaptionskip}{0cm}
    \setlength{\belowcaptionskip}{0cm}
    \begin{center}
    \setlength\tabcolsep{3pt}
    \begin{tabular}{cccccccc}
    \toprule    
    Pose$\_$X (m)  & 0.33 & 0.27 & 0.21  & 0.15  & 0.09  & 0.03 & -0.02 \\
    Mask ($\%$) & 64.3 & 58.7 & 51.9  & 43.2  & 33.2  & 17.7 & 14.9  \\
    \midrule
    Pose$\_$Y (m)  & 0.20 & 0.16 & 0.12  & 0.09  & 0.05  & 0.02 & -0.02  \\
    Mask ($\%$) & 28.9 & 24.2 & 19.6  & 15.3  & 11.2  & 7.1  & 7.6  \\
    \midrule
    Pose$\_$Z (m)  & 0.33 & 0.27 & 0.21  & 0.15  & 0.09  & 0.03 & -0.02 \\
    Mask ($\%$) & 62.2 & 56.7 & 50.0  & 41.9  & 31.5  & 16.7  & 16.0 \\ 
    \bottomrule

    \end{tabular}
    \end{center}
    \caption{The camera movement from the current pose along different axes
     and their corresponding hole-to-image area ratios, Mask ($\%$).
     Sign `-' indicates moving towards the negative direction of axis. }
    \label{tab:mask}
    \vspace{-.43cm}
\end{table}

\subsection{Panoramic 3D Gaussian Splatting}
\begin{figure}[t]
    \centering
    \subfloat[COLMAP Output]
    {
        \label{COLMAP_Ori}
        \includegraphics[width=0.45\linewidth]{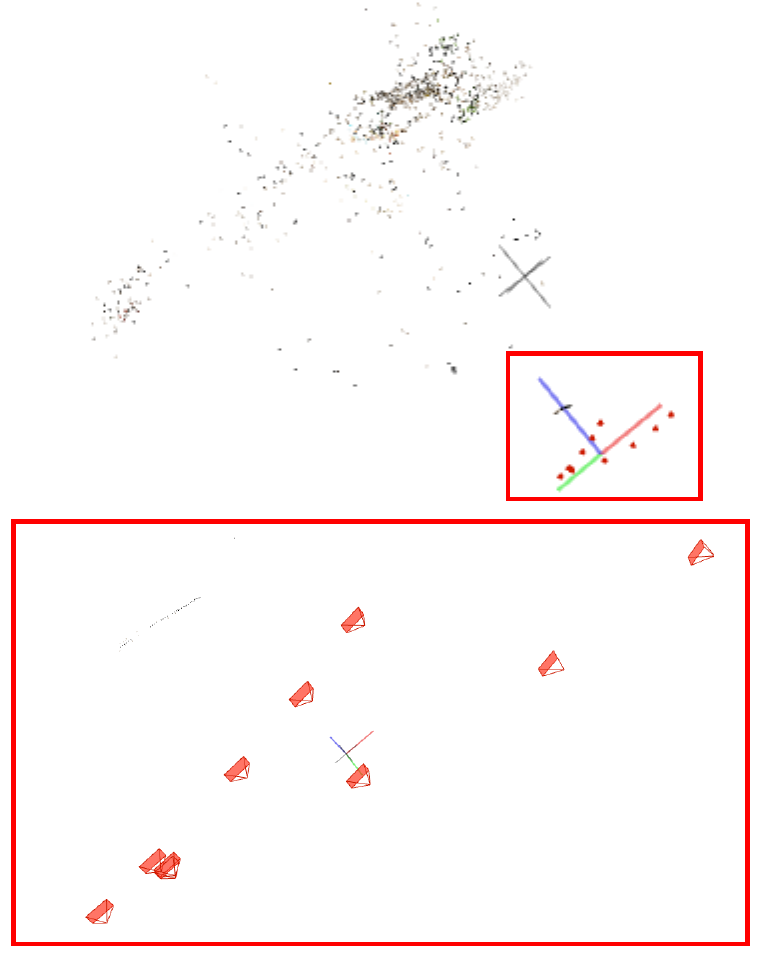}
    }\hspace{0.1cm}
    \subfloat[MVP + COLMAP (Ours)]
    {
    \label{COLMAP_Our}
        \includegraphics[width=0.45\linewidth]{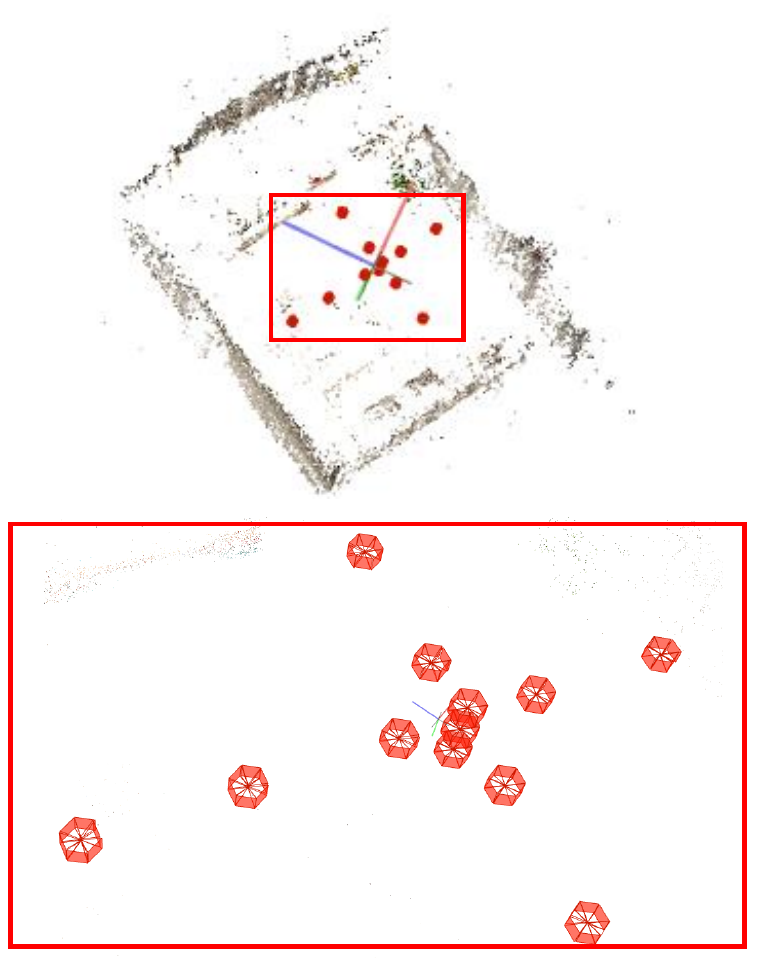}
    }
    \vspace{-0.2cm}
    \caption{The visual comparison of the original COLAMP output and our projection for panoramic input. 
    It is evident that our method is capable of obtaining accurate point clouds and camera poses.}
    \label{fig:colmap}
    \vspace{-0.3cm}
\end{figure}

The original inputs for 3DGS are multiple RGB perspective views. 
Following the COLMAP \cite{schonberger2016structure} pipeline, sparse point clouds and camera parameters are obtained.
Nevertheless, algorithms within COLMAP pertaining on perspective views exclusively 
exhibit inadequacies when confronted with panoramic perspectives, 
leading to a disorderly reconstruction outcome.
As shown in Figure \ref{COLMAP_Ori}, assuming the camera moves along $x, y, and z$ axes, 
the adoption of the original COLMAP fails to produce accurate point clouds and camera poses. 
This arises from the distinctive distortions and intricacies inherent in panoramas, 
making the application of conventional SFM arduous 
in the endeavor to align spatial information across diverse viewpoints.

Therefore, we introduce MVP to solve the aforementioned problem. 
Specifically, given the panorama $S$ with size $W \times H$ 
and the requirements for $n$ perspective images $(V_1, V_2, ..., V_n)$ with size $R \times R$.
Firstly, we calculate the rotation matrix $R_i$ for each camera. 
For each perspective view $V_i (1 \leq i \leq n)$, we define a projection mapping function $P(S, V_i)$, 
which maps the pixels of the panorama to the perspective view. 
By projecting the panoramic pixels $(m, q), (0 \leq m < W, 0 \leq q < H$),  
new projected coordinates $(j, k), (0 \leq j, k <R)$ for the perspective view can be obtained. 
Then, we send the multi-view perspective images to COLMAP to obtain the point clouds required by 3DGS. 
As shown in Figure \ref{COLMAP_Our}, for multi-view panoramic inputs, 
our method enables the generation of accurate point clouds and camera poses, 
thereby allowing for seamless processing using 3DGS.
The loss function $L$ is defined as the weighted sum of $L_1$ and $L_{D-SSIM}$ \cite{wang2004image}:
\begin{eqnarray}
    L = (1 - \lambda) L_1 + \lambda L_{D-SSIM} ,
\end{eqnarray}
we follow \cite{kerbl20233d} to set $ \lambda = 0.2$.

\section{Experiments}\label{Exper}
\subsection{Implementations Details}
We implement our method on PyTorch. 
We use the pre-trained Diffusion360 \cite{feng2023diffusion360} and EGformer \cite{yun2023egformer} for panorama generation and depth estimation, respectively. 
We retrain the AOT-GAN \cite{zeng2022aggregated} on our synthesized dataset for inpainting, described in Section \ref{dataset}. 
We choose CLIP Score \cite{hessel2021clipscore}, Natural Image Quality Evaluator (NIQE) \cite{mittal2012making}, 
and Blind/Referenceless Image Spatial Quality Evaluator (BRISQUE) \cite{mittal2012no} to evaluate the rendering quality in an unsupervised manner. 
It takes about 15 minutes to generate a complete scene on a single NVIDIA RTX A6000 GPU with 49G memory. 
Specifically, panorama generation takes 10 seconds, the PNVI process takes approximately 2 minutes, 
acquiring 3DGS training data requires around 3 minutes, and scene generation takes 10 minutes.

\subsection{Panoramic Inpainting Dataset}\label{dataset}
Due to the absence of panoramic datasets with our mask distribution, it is essential to generate a corresponding dataset. 
Specifically, we select the synthetic dataset Structured3D \cite{zheng2020structured3d}, 
which comprises 21k photorealistic panoramic scenes.  
We select 14k images with complete scenes that are more realistic. 
Subsequently, for each panorama, we generate 16 types of masks using equations (12) to (14), 
corresponding to eight movement directions on the coordinate axis, 
with two movement units of 0.02$m$ and 0.04$m$ for each direction. 
Then we perform E2C projection for each panorama and mask image. 
Finally, there are a total of 84k perspective RGB images and 1344k masks. 
After obtaining the dataset, we retrain AOT-GAN \cite{zeng2022aggregated}, with all training and testing sizes set as 512 $\times$ 512.

\subsection{Comparisons with Other Methods}
To validate the effectiveness of our method, 
we conduct quantitative and qualitative comparisons with previous indoor scene generation methods,
including Text2Room \cite{hoellein2023text2room}, Set-the-Scene \cite{cohen2023set}, 
and SceneScape \cite{fridman2023scenescape}. We render 30 images of each scene for evaluation. 

\textbf{Quantitative Comparison.}
By giving an identical text prompt input, we test the generative performance of different methods. 
Since conventional image quality assessment metrics, such as PSNR and SSIM, are not applicable to our task, 
we adopt unsupervised evaluation metrics. 

As reported in Table \ref{tab:com}, Text2Room performs modestly due to the lack of global consistency. 
SceneScape suffers from decreased image quality caused by accumulated errors during long-distance movements. 
Set-the-Scene exhibits limited perceptual performance due to its lower resolution and texture quality. 
On the contrary, our method not only achieves superior performance in terms of CLIP Score, NIQE, and BRISQUE metrics, 
but also demonstrates the fastest generation speed. 
A fast generation process is important, since it is an obvious advantage for user-friendly tasks.

\begin{figure*}[htbp]
    \centering

    \includegraphics[width=0.99\textwidth]{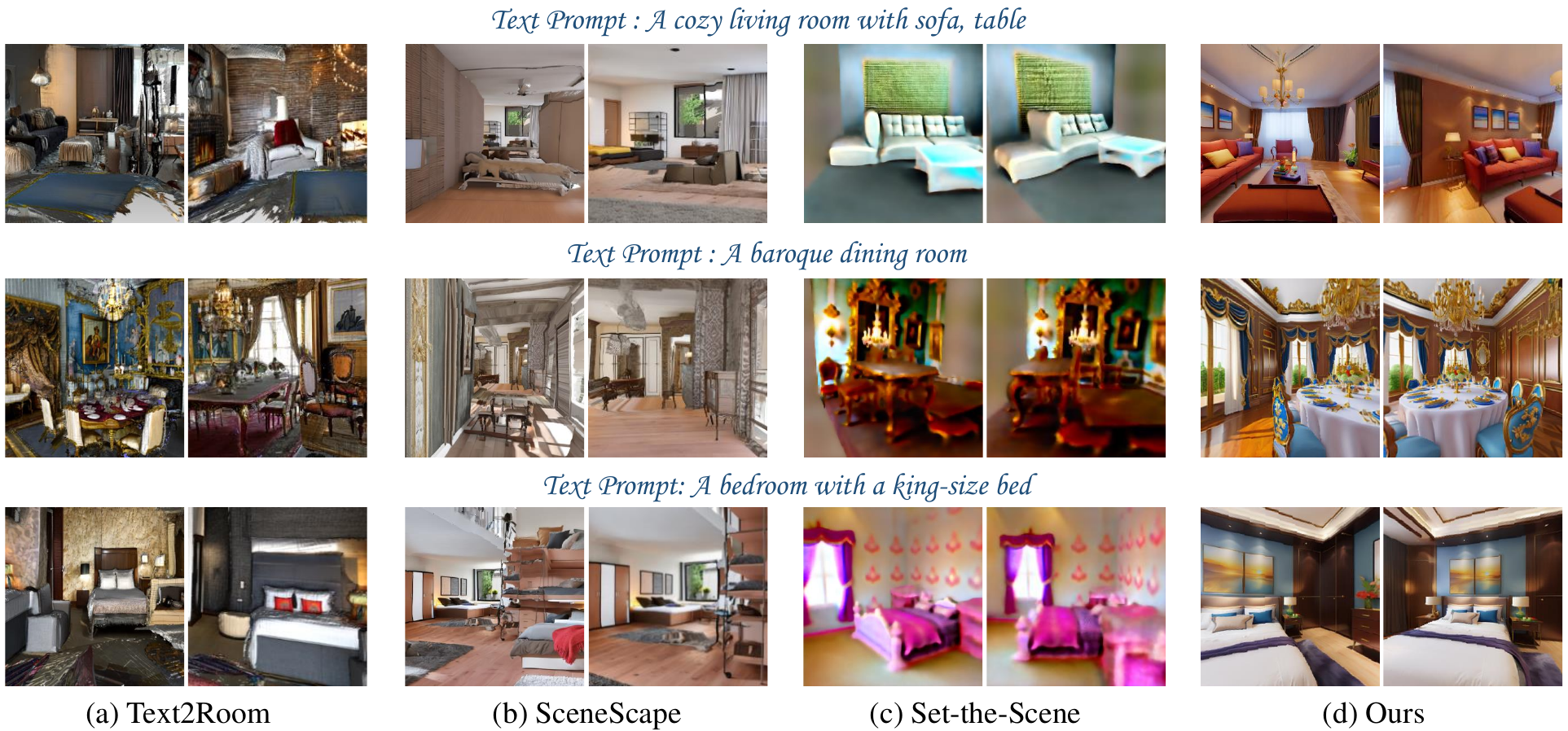}\vspace{0.2cm}
    
    \vspace{-0.3cm}
    \caption{Qualitative comparisons with other methods. 
        For each methods, we show the rendering views for the 1st and 5th frames. 
        Our method generats high-quality scenes from the same text prompts, while maintaining the scene consistency well.}
    \label{qua_com}
    \vspace{-0.3cm}
\end{figure*}

\textbf{Qualitative Comparison.}
Furthermore, to comprehensively validate the performance of our FastScene, 
we present the qualitative comparison results with other scene generation methods. 
We provide the same text prompt, such as common indoor scenes: bedroom, living room, dining room, etc., 
and then obtain the generation results of different methods. 
As shown in Figure \ref{qua_com}, Text2Room \cite{hoellein2023text2room} can generate faithful local views, 
but it fails to ensure consistency across the entire scene. 
SceneScape \cite{fridman2023scenescape} has the ability to generate long-range immersive views.
However, as the distance increases, 
the accumulation of errors results in a detrimental loss of details. 
Set-the-Scene \cite{cohen2023set} possesses the ability to generate editable scenes. 
However, its rendered images are blurry and texture quality is inadequate to meet perceptual needs. 
In comparison, our method generates high-quality scenes in a fast way,
and ensures the scene consistency as well. More scene generation results can be found in our \textbf{Supplementary Material}. 
\begin{table}[htbp]        
    \begin{center}
        \setlength\tabcolsep{1pt}
        \begin{tabular}{ccccc}
        \toprule
        Methods        & CLIP $\uparrow$ & NIQE $\downarrow$ & BRISQUE $\downarrow$ & Time(min/scene) \\
        \midrule
        Text2Room      & 28.1 & 5.4 & 28.4 & 70 \\
        Set-the-Scene  & 23.8 & 9.3 & 51.6 & 155 \\
        SceneScape     & 24.7 & 4.4 & 32.3 & 110 \\
        Ours           & \textbf{29.0} & \textbf{3.9} & \textbf{20.6} & \textbf{15}  \\
        \bottomrule
        \end{tabular}
        \end{center}
        \vspace{-.2cm}
        \caption{Quantitative comparison with other methods, 
        with all results tested on the same hardware device.}
        \label{tab:com}
        \vspace{-.3cm}
\end{table}

In conclusion, both quantitative and qualitative comparison experiments confirm that 
our method can rapidly and effectively generate globally consistent scenes with high-quality. 

\subsection{Extension Experiments on Panoramic Datasets}
\begin{figure}[b]
    \vspace{-0.2cm}
    \centering
    \subfloat[Matterport3D, 1K]
    {
        \includegraphics[width=0.48\linewidth]{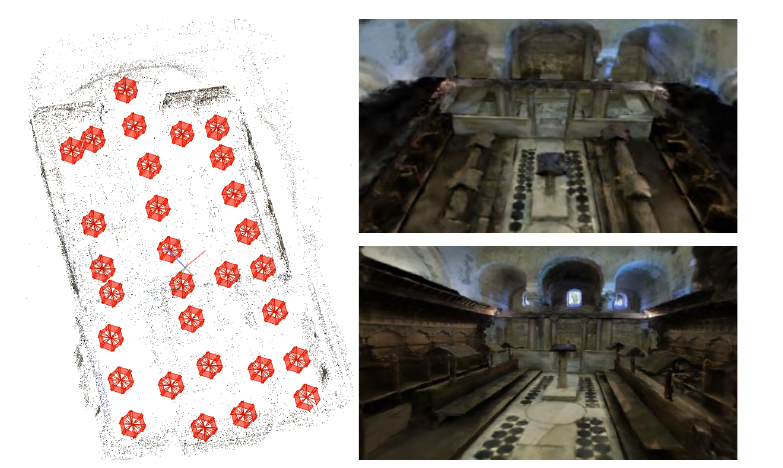}
    }
    \subfloat[Replica360, 4K]
    {
        \includegraphics[width=0.48\linewidth]{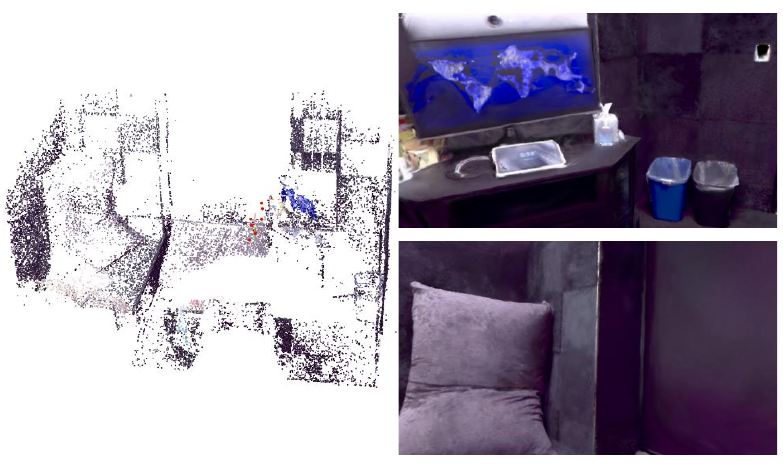}
    }
    \caption{The reconstruction results of indoor panoramic datasets, 
    validating that our method can effectively transfer to 360\textdegree \ datasets.}
    \label{exp_com2k}
    \vspace{-0.4cm}
\end{figure}

To validate the adaptability of our PNVI and MVP on existing panoramas for 3DGS, 
we conduct extension experiments on the Matterport3D 1k, 2k \cite{chang2017matterport3d}, and Replica360 4K \cite{straub2019replica} datasets, 
containing panoramas at resolutions of 1K, 2K, and 4K, respectively. 
As shown in Figure \ref{exp_com2k}, our method is capable of reconstructing 3D scenes from panoramas at different resolutions. 

\begin{figure*}[htbp]
    \centering
    \includegraphics[width=0.99\textwidth]{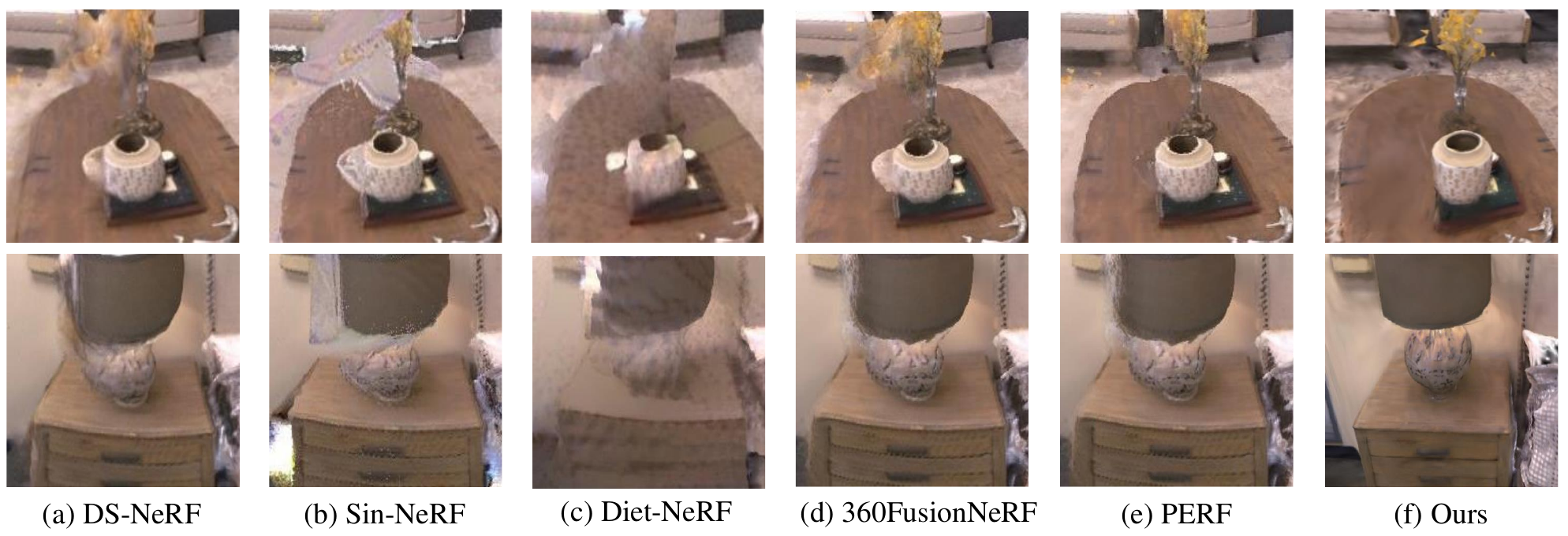}
    \caption{The extension experiments for existing panoramic datasets. 
    It is clear that our rendering quality surpasses other methods, 
    validating our FastScene can effectively transfer to different panoramas.}
    \label{exp_com4k}
    \vspace{-0.2cm}
\end{figure*}

Furthermore, to demonstrate the effectiveness of our method, we compare the performance with panoramic novel views synthesis works on Replica360 4K: 
DS-NeRF \cite{deng2022depth}, SinNeRF \cite{xu2022sinnerf}, DietNeRF \cite{jain2021putting}, 
360FusionNeRF \cite{kulkarni2023360fusionnerf}, and PERF \cite{wang2023perf}.
These NeRF-based methods inherently lack the ability to infer occluded content and have insufficient geometric constraints for panoramic structures. 
As a result, they suffer from varying degrees of blurriness and reduced quality, as shown in Figure \ref{exp_com4k} and Table \ref{tab:true}. 
Among them, PERF exhibits relatively satisfactory results, 
but it lacks consideration of panoramic geometric information, and there is a certain degree of quality degradation. 
On the contrary, 
we design PNVI and MVP to fully consider the constraints of the panoramic structure, while employing 3DGS rather than NeRF architecture, 
resulting in higher rendering quality in both quantitative and qualitative performance. 

The extension experiments further demonstrate that our method can be extended to existing panoramas and perform high-quality novel view synthesis. 

\begin{table}[b]
    \vspace{-.2cm}        
    \begin{center}
    \setlength\tabcolsep{8pt}
    \begin{tabular}{cccc}
    \toprule
    Methods         & PSNR $\uparrow$ & SSIM $\uparrow$ & LPIPS $\downarrow$ \\
    \midrule
    DS-NeRF         & 23.29 & 0.834 & 0.265 \\
    SinNeRF         & 22.70 & 0.826 & 0.251 \\
    DietNeRF        & 23.24 & 0.836 & 0.291 \\
    360FusionNeRF   & 21.54 & 0.833 & 0.245 \\
    PERF            & 23.49 & 0.838 & \textbf{0.244} \\
    Ours            & \textbf{23.52} & \textbf{0.841} & 0.245 \\
    \bottomrule
    \end{tabular}
    \end{center}
    \vspace{-.3cm}
    \caption{Quantitative comparisons on Replica360 dataset. 
    Our FastScene achieves better quantitative evaluation results 
    than other views rendering methods.}
    \label{tab:true}
    \vspace{-.4cm}
\end{table}

\subsection{Ablation Studies}
To validate the necessity of our inpainting mode 
and the effectiveness of the progressive inpainting strategy in PNVI,  
we design two corresponding ablation studies for different inpainting modes: 

\textbf{Directly inpainting panorama.}
We first retrain the AOT-GAN on our synthesized panoramic dataset, and then directly perform inpainting on panoramas. 
We find the performance is ideal for small-distance movements, as shown in Figure 8(a). 

However, as the movement distance increases, noticeable distortion and edge-blurring artifacts appear, as shown in Figure 8(b). 
This is due to the accumulated errors in depth estimation and the projection errors in the inpainting process. 
Additionally, due to discrepancies between the truth depth values in the dataset and our estimations, 
the distribution of holes is not entirely consistent between the training and inference stages. 

\textbf{Inpainting for a large distance.}
To validate the effectiveness of our progressive inpainting strategy, we perform inpainting on novel views with large camera poses, 
rather than incrementally moving. 
According to Figure 8(c), it is evident that directly inpainting large poses results in serious artifacts, 
which affects subsequent processing. 
When there is a large hole-to-image ratio, it becomes challenging to ensure the generation quality, 
thus affecting the consistency of the overall scene. 
By progressively inpainting the cubemap images, our PNVI strategy can address distortion and edge-blurring issues, 
as shown in Figure 8(a).

Table \ref{tab:aba} reports the quantitative comparisons of different inpainting modes, 
where our FastScene achieves the best performance in scene generation. 
In summary, the ablation studies further demonstrate the effectiveness of our method.
\begin{figure}[!t]
    \vspace{-0.15cm}
    \centering
    \includegraphics[width=0.48\textwidth]{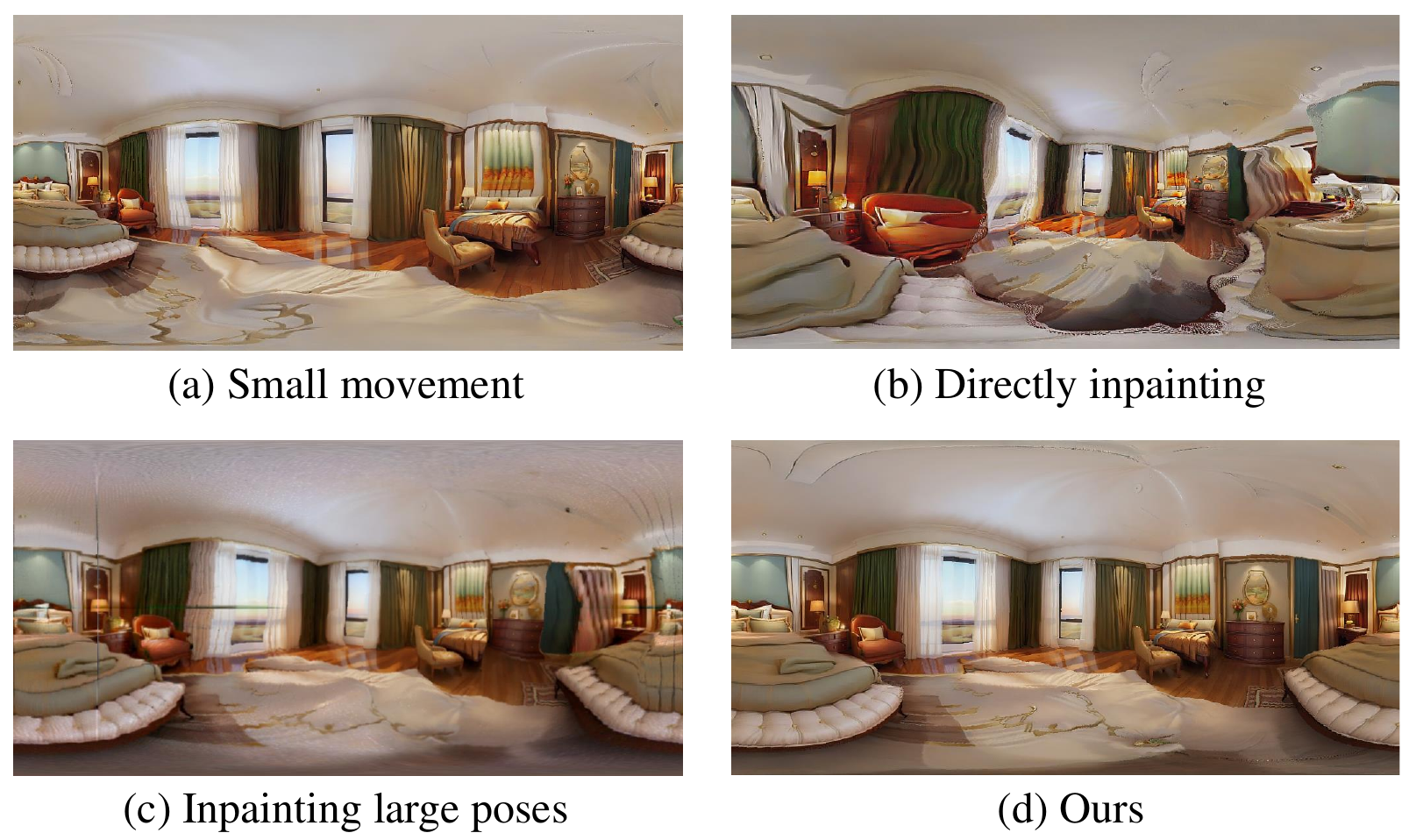}
    
    \vspace{-0.2cm}
    \caption{Different inpainting modes. Directly inpainting results in distortion and edge blurring, 
    and inpainting at large poses leads to content artifacts. 
    Our cubemap inpainting addresses these issues.}
    \label{Pano_niuqu}
    \vspace{-0.25cm}
\end{figure}
\begin{table}[htbp]        
\begin{center}
    \setlength\tabcolsep{1.35pt}
    \begin{tabular}{ccccc}
    \toprule
    Methods        & CLIP $\uparrow$ & NIQE $\downarrow$ & BRISQUE $\downarrow$ & TIME(min) \\
    \midrule
    Directly        & 27.3 & 6.8 & 45.1 & 14 \\
    Large-distance  & 25.7 & 7.4 & 42.3 & \textbf{11} \\
    Ours            & \textbf{29.0} & \textbf{3.9} & \textbf{20.6} & 15 \\ 
    \bottomrule
    \end{tabular}
    \end{center}
    \vspace{-.4cm}
    \caption{Ablation studies for directly, large-distance, and cubemap inpainting.
    We retrain AOT-GAN on our synthetic dataset.}
    \label{tab:aba}
    \vspace{-.4cm}
\end{table}

\section{Conclusion}\label{conc}
We propose a fast Text-to-3D indoor scene generation framework FastScene, 
exhibiting satisfactory scene quality and consistency. 
For users, FastScene only requires a text prompt without designing motion parameters, 
and provide a complete high-quality 3D scene in only 15 minutes. 
The proposed PNVI with CVS can generate consistent novel panoramic views, 
while MVP projects them into perspective views, facilitating 3DGS reconstruction. 
Extensive experiments demonstrate the effectiveness of our method. 
FastScene provides a user-friendly scene generation paradigm, and we believe it has wide-ranging potential applications.
In future work, we will focus on 3D scene editing and multimodal learning.

\appendix
\section*{Acknowledgments}
This work was supported by the National Natural Science Foundation of China (No.62071500), 
and Shenzhen Science and Technology Program (Grant No. JCYJ20230807111107015).
\bibliographystyle{named}
\bibliography{ijcai24}

\begin{thebibliography}{}

\bibitem[\protect\citeauthoryear{Barron \bgroup \em et al.\egroup
  }{2021}]{barron2021mip}
Jonathan~T Barron, Ben Mildenhall, Matthew Tancik, Peter Hedman, Ricardo
  Martin-Brualla, and Pratul~P Srinivasan.
\newblock Mip-nerf: A multiscale representation for anti-aliasing neural
  radiance fields.
\newblock In {\em Proceedings of the IEEE/CVF International Conference on
  Computer Vision}, pages 5855--5864, 2021.

\bibitem[\protect\citeauthoryear{Barron \bgroup \em et al.\egroup
  }{2022}]{barron2022mip}
Jonathan~T Barron, Ben Mildenhall, Dor Verbin, Pratul~P Srinivasan, and Peter
  Hedman.
\newblock Mip-nerf 360: Unbounded anti-aliased neural radiance fields.
\newblock In {\em Proceedings of the IEEE/CVF Conference on Computer Vision and
  Pattern Recognition}, pages 5470--5479, 2022.

\bibitem[\protect\citeauthoryear{Chang \bgroup \em et al.\egroup
  }{2017}]{chang2017matterport3d}
Angel Chang, Angela Dai, Thomas Funkhouser, Maciej Halber, Matthias Niebner,
  Manolis Savva, Shuran Song, Andy Zeng, and Yinda Zhang.
\newblock Matterport3d: Learning from rgb-d data in indoor environments.
\newblock In {\em International Conference on 3D Vision}, 2017.

\bibitem[\protect\citeauthoryear{Chen and Williams}{1993}]{chen1993view}
Shenchang~Eric Chen and Lance Williams.
\newblock View interpolation for image synthesis.
\newblock In {\em Proceedings of the 20th Annual Conference on Computer
  Graphics and Interactive Techniques}, pages 279--288, 1993.

\bibitem[\protect\citeauthoryear{Chen \bgroup \em et al.\egroup
  }{2022}]{chen2022text2light}
Zhaoxi Chen, Guangcong Wang, and Ziwei Liu.
\newblock Text2light: Zero-shot text-driven hdr panorama generation.
\newblock {\em ACM Transactions on Graphics}, 41(6):1--16, 2022.

\bibitem[\protect\citeauthoryear{Chen \bgroup \em et al.\egroup
  }{2023}]{chen2023testnerf}
Jiafu Chen, Boyan Ji, Zhanjie Zhang, Tianyi Chu, Zhiwen Zuo, Lei Zhao, Wei
  Xing, and Dongming Lu.
\newblock Testnerf: text-driven 3d style transfer via cross-modal learning.
\newblock In {\em Proceedings of the Thirty-Second International Joint
  Conference on Artificial Intelligence}, pages 5788--5796, 2023.

\bibitem[\protect\citeauthoryear{Chen \bgroup \em et al.\egroup
  }{2024}]{chen2024panogrf}
Zheng Chen, Yan-Pei Cao, Yuan-Chen Guo, Chen Wang, Ying Shan, and Song-Hai
  Zhang.
\newblock Panogrf: Generalizable spherical radiance fields for wide-baseline
  panoramas.
\newblock {\em Advances in Neural Information Processing Systems}, 36, 2024.

\bibitem[\protect\citeauthoryear{Cohen-Bar \bgroup \em et al.\egroup
  }{2023}]{cohen2023set}
Dana Cohen-Bar, Elad Richardson, Gal Metzer, Raja Giryes, and Daniel Cohen-Or.
\newblock Set-the-scene: Global-local training for generating controllable nerf
  scenes.
\newblock In {\em Proceedings of the IEEE/CVF International Conference on
  Computer Vision}, pages 2920--2929, 2023.

\bibitem[\protect\citeauthoryear{Debevec \bgroup \em et al.\egroup
  }{1996}]{debevec1996modeling}
Paul~E. Debevec, Camillo~J. Taylor, and Jitendra Malik.
\newblock Modeling and rendering architecture from photographs: A hybrid
  geometry- and image-based approach.
\newblock In {\em Proceedings of the 23rd Annual Conference on Computer
  Graphics and Interactive Techniques}, pages 11--20, 1996.

\bibitem[\protect\citeauthoryear{Deng \bgroup \em et al.\egroup
  }{2022}]{deng2022depth}
Kangle Deng, Andrew Liu, Jun-Yan Zhu, and Deva Ramanan.
\newblock Depth-supervised nerf: Fewer views and faster training for free.
\newblock In {\em Proceedings of the IEEE/CVF Conference on Computer Vision and
  Pattern Recognition}, pages 12882--12891, 2022.

\bibitem[\protect\citeauthoryear{Fang \bgroup \em et al.\egroup
  }{2023}]{fang2023ctrl}
Chuan Fang, Xiaotao Hu, Kunming Luo, and Ping Tan.
\newblock Ctrl-room: Controllable text-to-3d room meshes generation with layout
  constraints.
\newblock {\em arXiv preprint arXiv:2310.03602}, 2023.

\bibitem[\protect\citeauthoryear{Feng \bgroup \em et al.\egroup
  }{2023}]{feng2023diffusion360}
Mengyang Feng, Jinlin Liu, Miaomiao Cui, and Xuansong Xie.
\newblock Diffusion360: Seamless 360 degree panoramic image generation based on
  diffusion models.
\newblock {\em arXiv preprint arXiv:2311.13141}, 2023.

\bibitem[\protect\citeauthoryear{Fridman \bgroup \em et al.\egroup
  }{2024}]{fridman2023scenescape}
Rafail Fridman, Amit Abecasis, Yoni Kasten, and Tali Dekel.
\newblock Scenescape: Text-driven consistent scene generation.
\newblock {\em Advances in Neural Information Processing Systems}, 36, 2024.

\bibitem[\protect\citeauthoryear{Gu \bgroup \em et al.\egroup
  }{2023}]{gu2023nerfdiff}
Jiatao Gu, Alex Trevithick, Kai-En Lin, Joshua~M Susskind, Christian Theobalt,
  Lingjie Liu, and Ravi Ramamoorthi.
\newblock Nerfdiff: Single-image view synthesis with nerf-guided distillation
  from 3d-aware diffusion.
\newblock In {\em International Conference on Machine Learning}, pages
  11808--11826. PMLR, 2023.

\bibitem[\protect\citeauthoryear{Hessel \bgroup \em et al.\egroup
  }{2021}]{hessel2021clipscore}
Jack Hessel, Ari Holtzman, Maxwell Forbes, Ronan Le~Bras, and Yejin Choi.
\newblock Clipscore: A reference-free evaluation metric for image captioning.
\newblock In {\em Proceedings of the 2021 Conference on Empirical Methods in
  Natural Language Processing}, pages 7514--7528, 2021.

\bibitem[\protect\citeauthoryear{H\"ollein \bgroup \em et al.\egroup
  }{2023}]{hoellein2023text2room}
Lukas H\"ollein, Ang Cao, Andrew Owens, Justin Johnson, and Matthias
  Nie{\ss}ner.
\newblock Text2room: Extracting textured 3d meshes from 2d text-to-image
  models.
\newblock In {\em Proceedings of the IEEE/CVF International Conference on
  Computer Vision}, pages 7909--7920, October 2023.

\bibitem[\protect\citeauthoryear{Jain \bgroup \em et al.\egroup
  }{2021}]{jain2021putting}
Ajay Jain, Matthew Tancik, and Pieter Abbeel.
\newblock Putting nerf on a diet: Semantically consistent few-shot view
  synthesis.
\newblock In {\em Proceedings of the IEEE/CVF International Conference on
  Computer Vision}, pages 5885--5894, 2021.

\bibitem[\protect\citeauthoryear{Kazhdan \bgroup \em et al.\egroup
  }{2006}]{kazhdan2006poisson}
Michael Kazhdan, Matthew Bolitho, and Hugues Hoppe.
\newblock Poisson surface reconstruction.
\newblock In {\em Proceedings of the fourth Eurographics symposium on Geometry
  processing}, volume~7, page~0, 2006.

\bibitem[\protect\citeauthoryear{Kerbl \bgroup \em et al.\egroup
  }{2023}]{kerbl20233d}
Bernhard Kerbl, Georgios Kopanas, Thomas Leimk{\"u}hler, and George Drettakis.
\newblock 3d gaussian splatting for real-time radiance field rendering.
\newblock {\em ACM Transactions on Graphics}, 42(4):1--14, 2023.

\bibitem[\protect\citeauthoryear{Kulkarni \bgroup \em et al.\egroup
  }{2023}]{kulkarni2023360fusionnerf}
Shreyas Kulkarni, Peng Yin, and Sebastian Scherer.
\newblock 360fusionnerf: Panoramic neural radiance fields with joint guidance.
\newblock In {\em 2023 IEEE/RSJ International Conference on Intelligent Robots
  and Systems}, pages 7202--7209. IEEE, 2023.

\bibitem[\protect\citeauthoryear{Lin \bgroup \em et al.\egroup
  }{2023}]{lin2023magic3d}
Chen-Hsuan Lin, Jun Gao, Luming Tang, Towaki Takikawa, Xiaohui Zeng, Xun Huang,
  Karsten Kreis, Sanja Fidler, Ming-Yu Liu, and Tsung-Yi Lin.
\newblock Magic3d: High-resolution text-to-3d content creation.
\newblock In {\em Proceedings of the IEEE/CVF Conference on Computer Vision and
  Pattern Recognition}, pages 300--309, 2023.

\bibitem[\protect\citeauthoryear{Mildenhall \bgroup \em et al.\egroup
  }{2020}]{mildenhall2020nerf}
Ben Mildenhall, Pratul~P Srinivasan, Matthew Tancik, Jonathan~T Barron, Ravi
  Ramamoorthi, and Ren Ng.
\newblock Nerf: Representing scenes as neural radiance fields for view
  synthesis.
\newblock In {\em European Conference on Computer Vision}, pages 405--421.
  Springer, 2020.

\bibitem[\protect\citeauthoryear{Mirzaei \bgroup \em et al.\egroup
  }{2023}]{mirzaei2023reference}
Ashkan Mirzaei, Tristan Aumentado-Armstrong, Marcus~A. Brubaker, Jonathan
  Kelly, Alex Levinshtein, Konstantinos~G. Derpanis, and Igor Gilitschenski.
\newblock Reference-guided controllable inpainting of neural radiance fields.
\newblock In {\em Proceedings of the IEEE/CVF International Conference on
  Computer Vision}, 2023.

\bibitem[\protect\citeauthoryear{Mittal \bgroup \em et al.\egroup
  }{2012a}]{mittal2012no}
Anish Mittal, Anush~Krishna Moorthy, and Alan~Conrad Bovik.
\newblock No-reference image quality assessment in the spatial domain.
\newblock {\em IEEE Transactions on Image Processing}, 21(12):4695--4708, 2012.

\bibitem[\protect\citeauthoryear{Mittal \bgroup \em et al.\egroup
  }{2012b}]{mittal2012making}
Anish Mittal, Rajiv Soundararajan, and Alan~C Bovik.
\newblock Making a “completely blind” image quality analyzer.
\newblock {\em IEEE Signal processing letters}, 20(3):209--212, 2012.

\bibitem[\protect\citeauthoryear{M{\"u}ller \bgroup \em et al.\egroup
  }{2022}]{muller2022instant}
Thomas M{\"u}ller, Alex Evans, Christoph Schied, and Alexander Keller.
\newblock Instant neural graphics primitives with a multiresolution hash
  encoding.
\newblock {\em ACM Transactions on Graphics}, 41(4):1--15, 2022.

\bibitem[\protect\citeauthoryear{Poole \bgroup \em et al.\egroup
  }{2022}]{poole2022dreamfusion}
Ben Poole, Ajay Jain, Jonathan~T Barron, and Ben Mildenhall.
\newblock Dreamfusion: Text-to-3d using 2d diffusion.
\newblock In {\em The Eleventh International Conference on Learning
  Representations}, 2022.

\bibitem[\protect\citeauthoryear{Rombach \bgroup \em et al.\egroup
  }{2022}]{rombach2022high}
Robin Rombach, Andreas Blattmann, Dominik Lorenz, Patrick Esser, and Bj{\"o}rn
  Ommer.
\newblock High-resolution image synthesis with latent diffusion models.
\newblock In {\em Proceedings of the IEEE/CVF conference on Computer Vision and
  Pattern Recognition}, pages 10684--10695, 2022.

\bibitem[\protect\citeauthoryear{Sajjadi \bgroup \em et al.\egroup
  }{2022}]{sajjadi2022scene}
Mehdi~SM Sajjadi, Henning Meyer, Etienne Pot, Urs Bergmann, Klaus Greff, Noha
  Radwan, Suhani Vora, Mario Lu{\v{c}}i{\'c}, Daniel Duckworth, Alexey
  Dosovitskiy, et~al.
\newblock Scene representation transformer: Geometry-free novel view synthesis
  through set-latent scene representations.
\newblock In {\em Proceedings of the IEEE/CVF Conference on Computer Vision and
  Pattern Recognition}, pages 6229--6238, 2022.

\bibitem[\protect\citeauthoryear{Schonberger and
  Frahm}{2016}]{schonberger2016structure}
Johannes~L Schonberger and Jan-Michael Frahm.
\newblock Structure-from-motion revisited.
\newblock In {\em Proceedings of the IEEE Conference on Computer Vision and
  Pattern Recognition}, pages 4104--4113, 2016.

\bibitem[\protect\citeauthoryear{Shen \bgroup \em et al.\egroup
  }{2023}]{shen2023anything}
Qiuhong Shen, Xingyi Yang, and Xinchao Wang.
\newblock Anything-3d: Towards single-view anything reconstruction in the wild.
\newblock {\em arXiv preprint arXiv:2304.10261}, 2023.

\bibitem[\protect\citeauthoryear{Snavely \bgroup \em et al.\egroup
  }{2006}]{snavely2006photo}
Noah Snavely, Steven~M Seitz, and Richard Szeliski.
\newblock Photo tourism: exploring photo collections in 3d.
\newblock In {\em ACM SIGGRAPH}, pages 835--846, 2006.

\bibitem[\protect\citeauthoryear{Straub \bgroup \em et al.\egroup
  }{2019}]{straub2019replica}
Julian Straub, Thomas Whelan, Lingni Ma, Yufan Chen, Erik Wijmans, Simon Green,
  Jakob~J Engel, Raul Mur-Artal, Carl Ren, Shobhit Verma, et~al.
\newblock The replica dataset: A digital replica of indoor spaces.
\newblock {\em arXiv preprint arXiv:1906.05797}, 2019.

\bibitem[\protect\citeauthoryear{Tang \bgroup \em et al.\egroup
  }{2023}]{tang2023mvdiffusion}
Shitao Tang, Fuyang Zhang, Jiacheng Chen, Peng Wang, and Yasutaka Furukawa.
\newblock Mvdiffusion: Enabling holistic multi-view image generation with
  correspondence-aware diffusion.
\newblock {\em arXiv preprint arXiv:2307.01097}, 2023.

\bibitem[\protect\citeauthoryear{Waechter \bgroup \em et al.\egroup
  }{2014}]{waechter2014let}
Michael Waechter, Nils Moehrle, and Michael Goesele.
\newblock Let there be color! large-scale texturing of 3d reconstructions.
\newblock In {\em European Conference on Computer Vision}, pages 836--850.
  Springer, 2014.

\bibitem[\protect\citeauthoryear{Wang \bgroup \em et al.\egroup
  }{2004}]{wang2004image}
Zhou Wang, Alan~C Bovik, Hamid~R Sheikh, and Eero~P Simoncelli.
\newblock Image quality assessment: from error visibility to structural
  similarity.
\newblock {\em IEEE Transactions on Image Processing}, 13(4):600--612, 2004.

\bibitem[\protect\citeauthoryear{Wang \bgroup \em et al.\egroup
  }{2024a}]{wang2023perf}
Guangcong Wang, Peng Wang, Zhaoxi Chen, Wenping Wang, Chen~Change Loy, and
  Ziwei Liu.
\newblock Perf: Panoramic neural radiance field from a single panorama.
\newblock {\em IEEE Transactions on Pattern Analysis and Machine Intelligence},
  2024.

\bibitem[\protect\citeauthoryear{Wang \bgroup \em et al.\egroup
  }{2024b}]{wang2024customizing}
Hai Wang, Xiaoyu Xiang, Yuchen Fan, and Jing-Hao Xue.
\newblock Customizing 360-degree panoramas through text-to-image diffusion
  models.
\newblock In {\em Proceedings of the IEEE/CVF Winter Conference on Applications
  of Computer Vision}, pages 4933--4943, 2024.

\bibitem[\protect\citeauthoryear{Xu \bgroup \em et al.\egroup
  }{2022}]{xu2022sinnerf}
Dejia Xu, Yifan Jiang, Peihao Wang, Zhiwen Fan, Humphrey Shi, and Zhangyang
  Wang.
\newblock Sinnerf: Training neural radiance fields on complex scenes from a
  single image.
\newblock In {\em European Conference on Computer Vision}, pages 736--753.
  Springer, 2022.

\bibitem[\protect\citeauthoryear{Yun \bgroup \em et al.\egroup
  }{2023}]{yun2023egformer}
Ilwi Yun, Chanyong Shin, Hyunku Lee, Hyuk-Jae Lee, and Chae~Eun Rhee.
\newblock Egformer: Equirectangular geometry-biased transformer for 360 depth
  estimation.
\newblock In {\em Proceedings of the IEEE/CVF International Conference on
  Computer Vision}, pages 6101--6112, 2023.

\bibitem[\protect\citeauthoryear{Zeng \bgroup \em et al.\egroup
  }{2022}]{zeng2022aggregated}
Yanhong Zeng, Jianlong Fu, Hongyang Chao, and Baining Guo.
\newblock Aggregated contextual transformations for high-resolution image
  inpainting.
\newblock {\em IEEE Transactions on Visualization and Computer Graphics}, 2022.

\bibitem[\protect\citeauthoryear{Zhang \bgroup \em et al.\egroup
  }{2023}]{zhang2023adding}
Lvmin Zhang, Anyi Rao, and Maneesh Agrawala.
\newblock Adding conditional control to text-to-image diffusion models.
\newblock In {\em Proceedings of the IEEE/CVF International Conference on
  Computer Vision}, pages 3836--3847, 2023.

\bibitem[\protect\citeauthoryear{Zhang \bgroup \em et al.\egroup
  }{2024}]{zhang2024text2nerf}
Jingbo Zhang, Xiaoyu Li, Ziyu Wan, Can Wang, and Jing Liao.
\newblock Text2nerf: Text-driven 3d scene generation with neural radiance
  fields.
\newblock {\em IEEE Transactions on Visualization and Computer Graphics}, 2024.

\bibitem[\protect\citeauthoryear{Zheng \bgroup \em et al.\egroup
  }{2020}]{zheng2020structured3d}
Jia Zheng, Junfei Zhang, Jing Li, Rui Tang, Shenghua Gao, and Zihan Zhou.
\newblock Structured3d: A large photo-realistic dataset for structured 3d
  modeling.
\newblock In {\em European Conference on Computer Vision}, pages 519--535.
  Springer, 2020.

\end{thebibliography}

\end{document}